\documentclass[sigconf,nonacm]{acmart}

\usepackage{balance}
\usepackage{graphicx}
\usepackage{amsmath}
\usepackage{algorithmic}
\usepackage{algorithm}
\usepackage{url}
\usepackage{xspace}
\usepackage{hyperref}

\usepackage{tikz}
\usetikzlibrary{positioning,shapes,arrows.meta,fit,backgrounds,automata,calc,shapes.multipart,decorations.pathreplacing}

\newcommand{\gaia}{\textsc{Gaia}\xspace}
\newcommand{\tci}{\textit{TCI}\xspace}
\newcommand{\stcc}{\textit{STCC}\xspace}

\title{GAIA: A General Agency Interaction Architecture for LLM-Human B2B Negotiation \& Screening}

\author{Siming Zhao}
\affiliation{%
  \institution{Alibaba.com US E-Commerce}
  \city{Sunnyvale}
  \state{CA}
  \country{USA}
}
\email{hengshou.zsm@alibaba-inc.com}

\author{Qi Li}
\affiliation{%
  \institution{Chinese University of Hong Kong, Shenzhen}
  \city{Shenzhen}
  \country{China}
}
\email{liqi@cuhk.edu.cn}

\begin{abstract}
Organizations are increasingly exploring delegation of screening and negotiation tasks to AI systems, yet deployment in high-stakes B2B settings is constrained by governance: preventing unauthorized commitments, ensuring sufficient information before bargaining, and maintaining effective human oversight and auditability. Prior work on Large Language Model (LLM) negotiation largely emphasizes autonomous bargaining in agent-to-agent settings and omits practical requirements such as staged information gathering, explicit authorization boundaries, and systematic feedback integration. We propose \gaia, a governance-first framework for LLM-human agency in B2B negotiation and screening. \gaia{} defines three essential roles—Principal (human), Delegate (LLM agent), and Counterparty—with an optional Critic to enhance performance, and organizes interactions through three mechanisms: information-gated progression that separates screening from negotiation; dual feedback integration that merges AI critique with lightweight human corrections; and authorization boundaries with explicit escalation paths. Our contributions are fourfold: (1)~a formal governance framework with three coordinated mechanisms and four safety invariants for delegation with bounded authorization; (2)~information-gated progression via task-completeness tracking (\tci) and explicit state transitions that separate screening from commitment; (3)~dual feedback integration combining AI Critic suggestions with human oversight through parallel learning channels; and (4)~a hybrid validation blueprint that proposes automated protocol metrics combined with human judgment of outcomes and safety. By bridging theory and practice, \gaia{} offers a reproducible specification for safe, efficient, and accountable AI delegation that can be instantiated across procurement, real estate, and staffing workflows.

\end{abstract}


\begin{document}

\maketitle

% Main sections
\section{Introduction}
\label{sec:intro}

\subsection{Motivating Scenario}
A procurement team in a mid-size manufacturer delegates initial supplier screening and terms exploration to an AI Delegate while retaining ultimate authority over commitments. The team runs 20 parallel engagements to source a critical material across multiple regions and quality grades. Early messages often contain ambiguity about specification, volume bands, lead-time constraints, certification requirements, payment terms, and warranty coverage. Business policy imposes strict authorization boundaries: the Delegate may ask questions, summarize options, and draft non-binding proposals, but may not agree to prices, volumes, exclusivity, or binding language without approval. Failure modes include premature bargaining on incomplete information, missed safety issues (e.g., binding phrasing, privacy disclosures), and inefficient back-and-forth that overloads human reviewers.

\subsection{The Agency Challenge}
This scenario illustrates the broader problem of delegated decision-making under uncertainty. Effective agency requires: (i)~gathering sufficient information before bargaining; (ii)~progressing conversations only when information completeness meets predefined gates; and (iii)~maintaining human oversight with clear escalation when authorization boundaries are approached. Without structure, LLMs can negotiate confidently on partial context or drift into commitments~\cite{hacks2024,kwon2024effective}; purely human workflows, however, do not scale across many simultaneous negotiations. The design challenge is to balance autonomy with control: minimize rounds to readiness while preserving safety, auditability, and outcome quality.

\subsection{Gaps in Existing Approaches}
Prior work on LLM negotiation~\cite{kwon2024effective,negotiationarena2024,bargaining2024,fu2023improving} offers valuable building blocks but leaves key deployment needs unaddressed:
\begin{itemize}
    \item Focus on agent-to-agent bargaining~\cite{cooperation2024,fu2023improving} rather than LLM-\allowbreak{}human-\allowbreak{}LLM agency workflows.
    \item No explicit modeling of a screening phase separate from negotiation.
    \item Lack of an information completeness measure to gate progression (and prevent premature commitment).
    \item Limited formalization of authorization boundaries, escalation triggers, and audit trails despite known risks~\cite{automated-risky2024}.
    \item Ad-hoc architectures without reproducible protocol specifications, despite advances in multi-agent frameworks~\cite{wang2023survey,li2023camel}.
\end{itemize}

\subsection{Our Approach: GAIA Overview}
We propose \gaia, a governance-first framework for LLM-human agency in screening and negotiation. \gaia{} defines three essential roles—Principal (human), Delegate (LLM agent), and Counterparty—with an optional Critic to enhance performance. Interactions are organized through three mechanisms: (1)~information-gated progression that explicitly separates screening from negotiation; (2)~dual feedback integration that merges AI Critic suggestions with lightweight human corrections; and (3)~authorization boundaries with structured escalation and non-binding phrasing safeguards. A phased workflow advances through START $\rightarrow$ a high-value opening question (Single-Turn Constraint Clarification, \stcc) $\rightarrow$ SCREEN $\rightarrow$ NEGOTIATE $\rightarrow$ SUMMARIZE $\rightarrow$ terminal states \{AGREE, NO\_DEAL, ESCALATE, STALL\}. Information completeness is tracked via a task-specific checklist so that negotiation commences only when required fields are sufficiently revealed, while a safety-oversight layer preflights draft replies for binding language and boundary overruns. This governance-first design aims to make delegation auditable, efficient, and safe across domains.

\subsection{Key Contributions}
\begin{enumerate}
    \item \textbf{Governance Framework for AI Delegation.}
    We present three formal governance mechanisms—information-gated progression with task-completeness tracking (\tci), dual feedback integration combining AI Critic and human oversight, and authorization boundaries with structured escalation—demonstrating how they coordinate to prevent premature commitments while maintaining delegation efficiency. The framework is instantiated across three domains (procurement, real estate, staffing) to illustrate cross-domain applicability and provide concrete implementation patterns.
    
    \item \textbf{Formal Protocol Architecture with Safety Invariants.}
    We specify a typed negotiation state machine with commitment detection, bounded authorization checks, and escalation actions. We formalize four safety invariants—No-Unauthorized-Commitment, Information-Gate-Monotonicity, Escalation-Before-Irrevocable-Offer, and Traceability—and demonstrate protocol compliance via structured walkthroughs across three domain instantiations, providing a reproducible specification for safe agent delegation.
    
    \item \textbf{Hybrid Validation Blueprint.}
    We propose an evaluation methodology that mirrors \gaia's dual-channel architecture: automated protocol metrics (gate compliance, escalation rates, commitment detection accuracy) combined with human judgment of outcome quality (utility preservation, satisfaction, trust). This blueprint specifies concrete measurement approaches, proposed baselines, and statistical protocols to guide future empirical validation.
\end{enumerate}

\subsection{Paper Organization}
Section~\ref{sec:related} reviews related work on LLM negotiation and multi-agent systems. Section~\ref{sec:framework} presents the \gaia{} framework with formal definitions and protocol architecture. Section~\ref{sec:evaluation} outlines a hybrid evaluation approach combining automated metrics and human judgment. Section~\ref{sec:discussion} synthesizes key implications, acknowledges limitations, outlines future research directions, and concludes.

\section{Related Work}
\label{sec:related}

\subsection{LLM Negotiation and Bargaining}
Research on LLM negotiation has largely focused on agent-to-agent bargaining in stylized settings such as role-play, self-play, and benchmark platforms. Studies show that LLMs can produce contextually appropriate offers and concessions. Prompt structure and role conditioning materially affect outcomes~\cite{kwon2024effective,negotiationarena2024,bargaining2024}. Self-play and critique-then-revise paradigms improve tactical coherence and agreement rates~\cite{fu2023improving}. These approaches also reveal role-dependent capabilities (buyers vs.\ sellers), sensitivity to framing, and degradation under distribution shift. Recent work documents prompt vulnerabilities and reasoning gaps: models may accept unfavorable terms, fail to track constraints across turns, or inadvertently use binding language~\cite{hacks2024}.

\textbf{Gap.} Most evaluations assume complete or static information and begin bargaining immediately. They omit the information-gathering phase that precedes real negotiations. More critically, they lack governance mechanisms for enterprise deployment: no authorization boundaries that define what the agent may commit to, no escalation protocols for human review, and no auditable state transitions linking oversight decisions to outcomes. \gaia{} addresses this by introducing an explicit SCREEN phase with information-completeness gating and embedding commitment checks, authorization boundaries, and escalation triggers as first-class protocol transitions.

\subsection{Multi-Agent Architectures}
Multi-agent frameworks explore how role specialization, communication patterns, and coordination strategies shape emergent behavior. Role-based prompting, shared memories, and turn-taking protocols enable division of labor among agents~\cite{wang2023survey}. Systems such as CAMEL show that constrained roles and structured dialogues improve task focus and reduce drift~\cite{li2023camel}. Subsequent work investigates message passing, planning hierarchies, and tool-use orchestration to stabilize collaboration. For example, CAMEL uses inception prompting to assign complementary roles (AI assistant, AI user) and track dialogue history for consistency.

\textbf{Gap.} Many architectures remain ad-hoc and not production-aware. They lack formal state machines that specify valid transitions. They omit explicit gating criteria for phase progression—when should the system stop gathering information and start negotiating? They provide no standardized mechanisms for integrating human oversight signals. Consequently, these systems cannot provide the audit trails and authorization checks required in regulated environments. \gaia{} specifies a protocol with named states (SCREEN, NEGOTIATE, COMMIT), transition predicates (information-completeness thresholds, commitment detection), and human-in-the-loop integration points that produce auditable logs.

\subsection{Human-in-the-Loop Agent Systems}
Human-in-the-loop (HIL) systems study how to capture micro-feedback, approvals, and corrections during agent operation. Prior work shows that lightweight human inputs steer LLM behavior and improve adherence to preferences without expensive re-training~\cite{hil-software2024}. For instance, software development agents pause to request clarifications on ambiguous requirements or approval before committing code. In negotiation-adjacent settings, socially-aware agents use human guidance to balance rapport with task progress~\cite{assistive2024}. These systems typically collect feedback after actions, either for online correction or offline fine-tuning.

\textbf{Gap.} Three problems remain unaddressed. \emph{First}, there is no standardized protocol for \emph{when} to solicit human input: should the agent ask before every message, only when uncertain, or only when triggering policy violations? \emph{Second}, how should the system prioritize human corrections against automated critique when both are available but review budgets are limited? \emph{Third}, how should oversight decisions be logged to create audit trails that connect human approvals to negotiation outcomes? \gaia{} operationalizes HIL with explicit rules: human feedback has precedence over AI critique, both channels merge under token budgets, and escalation is mandatory when the agent detects potential unauthorized commitments.

\subsection{Safety and Authorization in Agentic AI}
Safety studies of agentic LLMs document risks including unauthorized commitments, privacy leaks, and susceptibility to adversarial tactics~\cite{wang2023survey,cooperation2024}. Guardrails often prove brittle under distributional shift: a classifier trained to detect commitment phrases may miss paraphrases or domain-specific terminology. Proposed mitigations include safety classifiers, commitment-phrase detection, rate limiting, and policy-aware rewriting. However, these are typically applied as post-hoc filters rather than integrated into the negotiation protocol. For example, a classifier might flag a message \emph{after} it is drafted but before sending. It does not guide the agent's planning or provide structured fallback (e.g., escalate to human, revise with constraints, or terminate).

\textbf{Gap.} Safety mechanisms should be protocol-native, not post-hoc patches. An integrated approach requires: (i)~\emph{commitment checks} that occur \emph{before} message generation, shaping what the agent considers; (ii)~\emph{authorization boundaries} encoded as constraints in the state machine, preventing transitions that exceed delegation scope; and (iii)~\emph{escalation triggers} that route edge cases to human review with full context. \gaia{} treats safety as stateful control flow: each outgoing message undergoes commitment detection against the principal's authorization scope, violations trigger escalation, and all decisions are logged for audit.

\subsection{Positioning and Gap Summary}
\gaia{} addresses five gaps in prior work. \textbf{First}, we model an explicit screening phase that precedes bargaining. Unlike benchmark studies that assume complete information, real negotiations require the agent to ask clarifying questions (e.g., ``What is your target delivery date?'' or ``Are there non-standard warranty requirements?'') and accumulate answers before making offers. \textbf{Second}, we introduce a \emph{Task-Completeness Index} (TCI) that measures information sufficiency and gates the transition from SCREEN to NEGOTIATE, preventing premature bargaining. \textbf{Third}, we unify automated critique (AI-generated feedback) and human micro-feedback in a single protocol with explicit precedence: human inputs override AI suggestions, both merge under context-window budgets. \textbf{Fourth}, we elevate safety and authorization to first-class protocol transitions. Each draft message undergoes commitment detection; violations trigger escalation with full context rather than silent rejection. \textbf{Fifth}, we provide a formal state machine with typed transitions, gating predicates, and audit logs—a reproducible specification rather than an ad-hoc agent recipe. Together, these contributions bridge the gap between benchmark performance and enterprise-grade, auditable agent deployment.

\section{The GAIA Framework}
\label{sec:framework}

\subsection{Reader's Map}
\gaia{} is a governance-first protocol for LLM-human agency in screening and negotiation. It defines three essential roles: Principal (human decision-maker), Delegate (LLM agent within bounded discretion), and Counterparty. Two optional components enhance the system: a Critic provides performance feedback, and a Moderator enforces safety and oversight. 

Interactions progress through a phased state machine with explicit gates. The flow is: START $\rightarrow$ high-value opening question (\stcc) $\rightarrow$ SCREEN $\rightarrow$ NEGOTIATE $\rightarrow$ SUMMARIZE $\rightarrow$ terminal states. Terminal states are: AGREE (successful deal), NO\_DEAL (impasse), ESCALATE (human review needed), or STALL (timeout). 

Three governance mechanisms structure the protocol. \textbf{First}, information-gated progression blocks bargaining until a Task-Completeness Index (\tci) threshold is met. \textbf{Second}, dual feedback integration merges Critic suggestions with human corrections under explicit precedence rules. \textbf{Third}, authorization boundaries trigger escalation when the agent approaches policy limits. 

\begin{figure}[tb]
  \centering
  \includegraphics[width=0.45\textwidth,keepaspectratio]{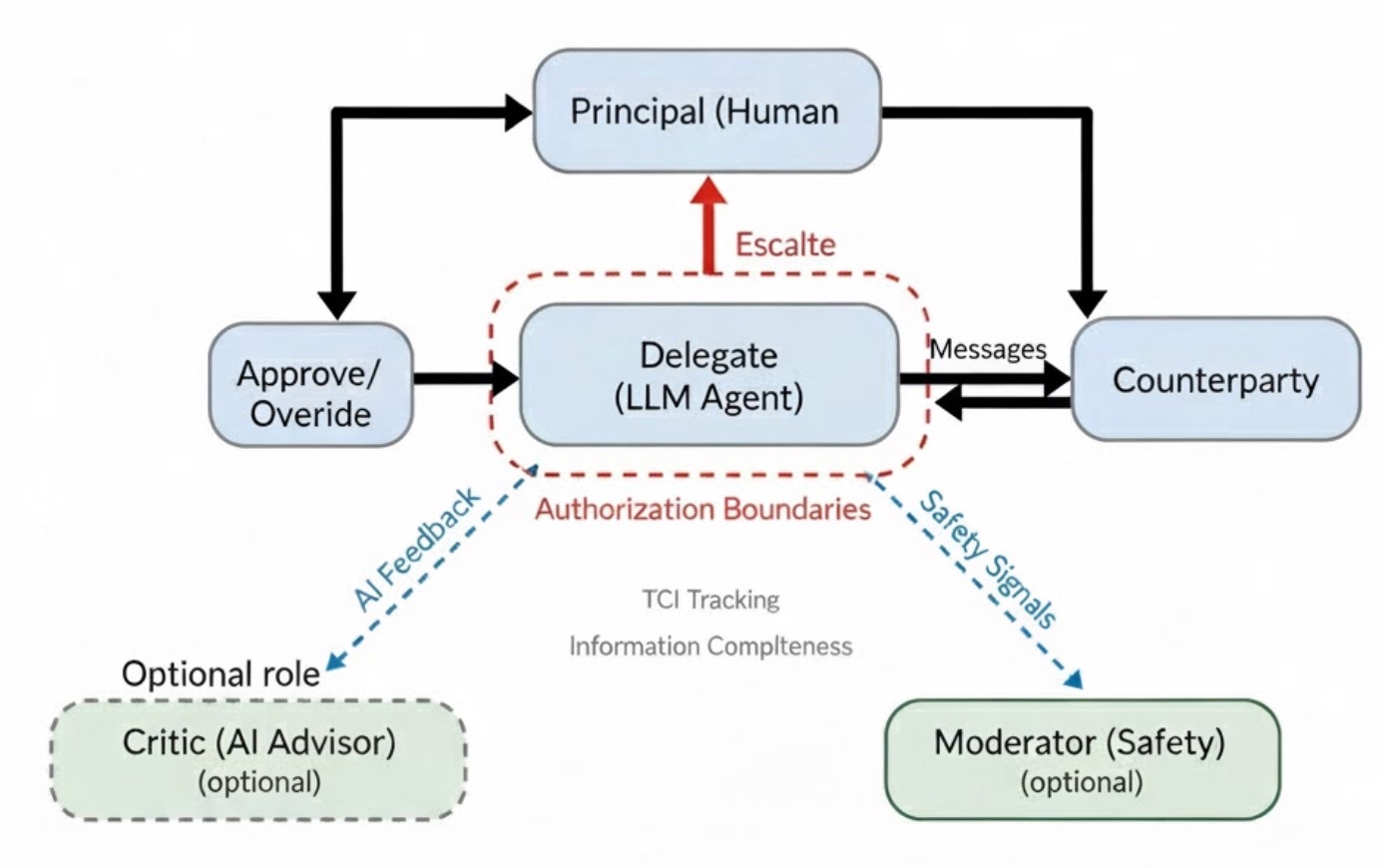}
  \caption{GAIA role architecture showing five roles (Principal, Delegate, Counterparty, Critic, Moderator) and their information flows. The Delegate operates within authorization boundaries, tracks TCI for information completeness, and escalates to the Principal on boundary violations or safety events.}
  \label{fig:role-architecture}
\end{figure}

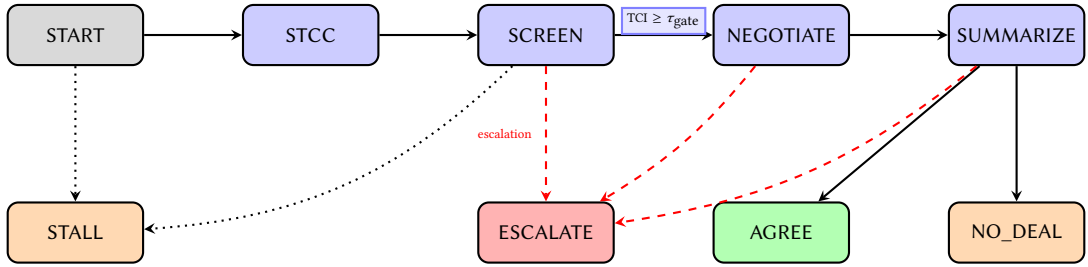
\begin{figure*}[tb]
  \centering
  \begin{tikzpicture}[
    node distance=1.2cm and 0.8cm,
    state/.style={rectangle, rounded corners, draw=black, thick, minimum width=1.8cm, minimum height=0.8cm, align=center, font=\small\sffamily},
    gray-state/.style={state, fill=gray!30},
    blue-state/.style={state, fill=blue!20},
    green-state/.style={state, fill=green!30},
    orange-state/.style={state, fill=orange!30},
    red-state/.style={state, fill=red!30},
    arrow/.style={->, >=stealth, thick},
    escalation/.style={->, >=stealth, thick, dashed, red},
    gate-label/.style={rectangle, draw=blue!50, fill=blue!10, font=\tiny, inner sep=2pt}
  ]
  
  % First row: main flow states (stretched for better alignment)
  \node[gray-state] (start) {START};
  \node[blue-state, right=1.3cm of start] (stcc) {STCC};
  \node[blue-state, right=1.3cm of stcc] (screen) {SCREEN};
  \node[blue-state, right=1.3cm of screen] (negotiate) {NEGOTIATE};
  \node[blue-state, right=1.3cm of negotiate] (summarize) {SUMMARIZE};
  
  % Second row: terminal states (all at same vertical level)
  \node[orange-state, below=1.8cm of start] (stall) {STALL};
  \node[red-state, below=1.8cm of screen] (escalate) {ESCALATE};
  \node[green-state, right=1.3cm of escalate] (agree) {AGREE};
  \node[orange-state, below=1.8cm of summarize] (nodeal) {NO\_DEAL};
  
  % Main flow arrows
  \draw[arrow] (start) -- (stcc);
  \draw[arrow] (stcc) -- (screen);
  \draw[arrow] (screen) -- node[above, gate-label] {$\text{TCI} \geq \tau_{\text{gate}}$} (negotiate);
  \draw[arrow] (negotiate) -- (summarize);
  \draw[arrow] (summarize) -- (agree);
  \draw[arrow] (summarize) -- (nodeal);
  
  % Escalation paths (dashed red arrows)
  \draw[escalation] (screen) -- node[left, font=\tiny, red, xshift=-2pt] {escalation} (escalate);
  \draw[escalation] (negotiate) to[bend left=10] (escalate);
  \draw[escalation] (summarize) to[bend left=15] (escalate);
  
  % Stall transitions (from any state)
  \draw[arrow, dotted] (start) -- (stall);
  \draw[arrow, dotted] (screen) to[bend left=20] (stall);
  
  \end{tikzpicture}
  \caption{GAIA state machine with information-gated transitions. States are color-coded: blue (active), green (terminal success), orange (terminal neutral), red (escalation). The TCI-based gate controls the SCREEN $\rightarrow$ NEGOTIATE transition, while escalation paths (dashed red) enable human intervention at any point.}
  \label{fig:state-machine}
\end{figure*}

\begin{figure}[tb]
  \centering
  \includegraphics[width=0.45\textwidth,keepaspectratio]{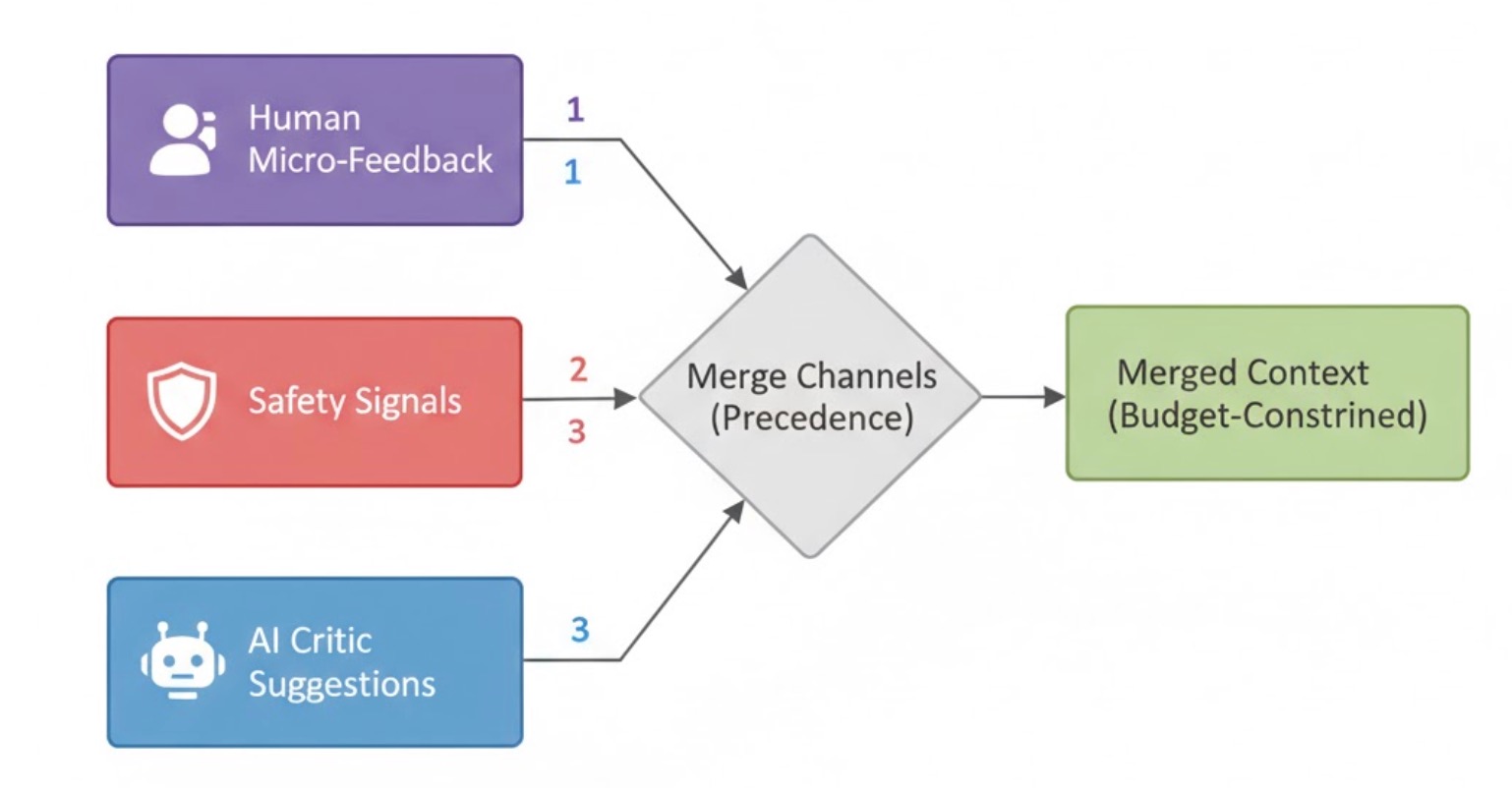}
  \caption{Dual learning channels coordination with precedence ordering. Three input streams (AI Critic, human micro-feedback, safety signals) are merged with explicit precedence: human overrides take highest priority, followed by safety signals, clarity tactics, and persuasion tactics. The merged context respects budget constraints.}
  \label{fig:dual-channels}
\end{figure}

Read this section as protocols: concise formal setup in~\ref{subsec:setup}, pseudocode-first architecture in~\ref{subsec:protocol}, practical patterns in~\ref{subsec:patterns}, and a staffing instantiation in~\ref{subsec:instantiation}. Figure callouts: Figure~\ref{fig:role-architecture} (role architecture overview), Figure~\ref{fig:state-machine} (state machine in~\ref{subsubsec:statemachine}), and Figure~\ref{fig:dual-channels} (dual channels coordination in~\ref{subsubsec:dualchannels}).

\subsection{Minimal Problem Setup}
\label{subsec:setup}
Definitions in this subsection use local numbering and are deliberately compact. After first use, we prefer natural language over acronym soup.

\begin{definition}[Agency Task]
An agency task is a tuple $(P, D, C, \Omega, \Phi, G)$ where $P$ is the Principal (human with preferences), $D$ the Delegate (LLM agent within bounded discretion), $C$ the Counterparty, $\Omega$ the outcome space, $\Phi$ the task constraints, and $G$ the guardrails.

\noindent\textit{Example:} In procurement, $P$ is the buyer, $D$ negotiates supplier terms, $C$ is the supplier, $\Omega$ includes price and delivery date, $\Phi$ specifies minimum quality standards, and $G$ prohibits the agent from committing to custom warranty clauses without approval.
\end{definition}

\begin{definition}[Screening vs Negotiation]
Screening reduces uncertainty about counterparty attributes via information elicitation. Negotiation optimizes outcome $\omega \in \Omega$ subject to $\Phi$ and $G$ once information is sufficiently complete.

\noindent\textit{Example:} In staffing, screening asks ``What is your work authorization status?'' and ``What time zones can you cover?'' to gather required information. Negotiation begins only after these answers are known, proposing compensation bands and start dates.
\end{definition}

\begin{definition}[Authorization Boundary]
The set of actions and commitments the Delegate may execute without escalation to the Principal. Boundaries are parameterized by thresholds (e.g., compensation bands, lead-time deviations) and prohibitions (binding language, PII disclosure).

\noindent\textit{Example:} A procurement agent may discuss prices within a \$50K–\$60K band but must escalate if the supplier requests \$62K. It may say ``We're exploring options'' but not ``We commit to ordering 500 units by Friday.''
\end{definition}

\begin{definition}[Task-Completeness Index, \tci]
$\text{TCI}(h) = \frac{|\text{revealed\_fields}(h)|}{|\text{required\_fields}|} \in [0, 1]$. Required fields are domain-specific. \tci{} is monotonic: it never decreases as information accumulates. \tci{} gates the SCREEN $\rightarrow$ NEGOTIATE transition.

\noindent\textit{Example:} In staffing, required fields are: work authorization, time-zone coverage, compensation range, start date, role title, and key skills (6 fields). If the agent has learned 4 of these, $\text{TCI} = 4/6 \approx 0.67$. If the gate threshold is $\tau_{\text{gate}} = 0.7$, the agent must ask at least one more clarifying question before proposing offers.
\end{definition}

\begin{definition}[Information Gain, IG]
$\text{IG}(t) = H(\theta | h_{t-1}) - H(\theta | h_t)$ measures how much a question reduces uncertainty. We use IG to evaluate question quality. Operationalization appears in Section~\ref{sec:evaluation}.

\noindent\textit{Example:} Asking ``What is your budget range: under \$50K, \$50K–\$100K, or over \$100K?'' reveals more information (higher IG) than asking ``Do you have a budget?'' which yields only yes/no.
\end{definition}

\begin{definition}[Screening Efficiency]
The number of dialogue rounds required to reach $\text{TCI} \geq \tau_{\text{complete}}$. Lower is better: efficient screening minimizes back-and-forth while gathering essential information.

\noindent\textit{Example:} If an agent reaches $\text{TCI} = 0.85$ in 3 rounds by asking targeted questions, it is more efficient than an agent that needs 7 rounds of vague inquiries to reach the same threshold.
\end{definition}

\noindent\textbf{Additional terminology:} \stcc{} (Single-Turn Constraint Clarification) is an early clarification pattern. It asks one carefully chosen, high-yield question to accelerate \tci{} growth. For example: ``Which factor most constrains this project: budget, timeline, or technical requirements?''

$\tau_{\text{gate}}$ is the minimum \tci{} needed to enter negotiation (typically 0.6–0.8). $\tau_{\text{complete}}$ is the threshold for readiness to commit (typically 0.7–0.9). In any delegated task, required\_fields define a checklist. \tci{} tracks progress and blocks bargaining until the checklist is sufficiently complete.

% \begin{figure}[tb]
%   \centering
%   \includegraphics[width=0.85\textwidth,keepaspectratio]{figures/figure4.png}
%   \caption{Three governance mechanisms in GAIA. \textbf{Left:} Information-gated progression uses TCI to control state transitions. \textbf{Center:} Dual feedback integration merges AI Critic, human, and safety inputs with explicit precedence. \textbf{Right:} Authorization boundaries with preflight checks and escalation paths ensure safe delegation.}
%   \label{fig:governance-mechanisms}
% \end{figure}

\subsection{Protocol Architecture}
\label{subsec:protocol}

\subsubsection{State Machine Specification}
\label{subsubsec:statemachine}

\noindent\textbf{States:} START $\rightarrow$ STCC $\rightarrow$ SCREEN $\rightarrow$ NEGOTIATE $\rightarrow$ SUMMARIZE $\rightarrow$ \{AGREE, NO\_DEAL, ESCALATE, STALL\}.

\noindent\textbf{Transition conditions:}
\begin{itemize}
    \item START $\rightarrow$ STCC: On session initiation.
    \item STCC $\rightarrow$ SCREEN: After \stcc response received.
    \item SCREEN $\rightarrow$ NEGOTIATE: When $\text{TCI} \geq \tau_{\text{gate}}$ and business logic permits.
    \item NEGOTIATE $\rightarrow$ SUMMARIZE: When agreement conditions approach or impasse detected.
    \item SUMMARIZE $\rightarrow$ AGREE: Commitments align within authorization.
    \item SUMMARIZE $\rightarrow$ NO\_DEAL: Impasse confirmed.
    \item Any $\rightarrow$ ESCALATE: Authorization boundary approached/violated, persistent ambiguity, or safety event.
    \item Any $\rightarrow$ STALL: Timeout/inactivity.
\end{itemize}

\noindent\textbf{Actions per transition:}
\begin{itemize}
    \item Persist state summaries and \tci{} ledger to memory.
    \item Notify Principal with a structured payload on ESCALATE.
    \item Re-enter SCREEN if \tci{} insufficient or Moderator confidence low.
    \item Run preflight commitment checks before any potentially binding phrasing.
\end{itemize}

\noindent\textbf{Formal description:} The protocol is a labeled transition system $M = (S, s_0, T, A)$ where:
\begin{itemize}
    \item $S = \{\text{START, STCC, SCREEN, NEGOTIATE, SUMMARIZE},$ \\ 
    $\text{AGREE, NO\_DEAL, ESCALATE, STALL}\}$
    \item $s_0 = \text{START}$
    \item $T \subseteq S \times A \times S$ encodes valid transitions
    \item $A$ includes Observable\-Actions $= \{\text{ask, summarize, propose},$ \\
    $\text{check\_safety, escalate, approve, decline}\}$
\end{itemize}

A gate function $g(h) = \mathbb{1}\{\text{TCI}(h) \geq \tau_{\text{gate}}\}$ controls the critical transition $\text{SCREEN} \rightarrow \text{NEGOTIATE}$. A completeness function $c(h) = \mathbb{1}\{\text{TCI}(h) \geq \tau_{\text{complete}}\}$ signals readiness to finalize. Safety constraints are enforced by a predicate $\text{Safe}(\text{draftReply}, \text{boundary})$. This predicate must return true before any message that could imply commitment.

\noindent\textit{Example:} If the agent drafts ``We can deliver 200 units by March 15,'' the safety predicate checks: (1)~Does this create a binding commitment? (2)~Is the quantity within authorization? (3)~Is the date within allowed windows? If any check fails, the system escalates or rewrites the draft.

\noindent\textbf{Invariants:}
\begin{itemize}
    \item \textbf{I1 (No premature bargaining):} For all histories $h$, if $\text{state} = \text{SCREEN}$ and $g(h) = 0$, then the agent cannot transition to NEGOTIATE. The agent must first gather more information to meet the gate threshold.
    
    \textit{Example:} If only 3 of 6 required fields are known ($\text{TCI} = 0.5$) and $\tau_{\text{gate}} = 0.7$, the agent cannot propose compensation offers. It must ask at least one more clarifying question.
    
    \item \textbf{I2 (Monotone information):} $\text{TCI}(h_t) \geq \text{TCI}(h_{t-1})$ by construction. Information is never forgotten. Therefore, the set of admissible transitions only expands, never shrinks.
    
    \textit{Example:} Once the agent learns the candidate's time-zone preference, this fact persists. The agent does not need to re-ask, and can proceed to propose meeting times.
    
    \item \textbf{I3 (Safety preflight):} Any message sent to the counterparty must satisfy $\text{Safe}(\text{draftReply}, \text{boundary})$ or trigger escalation. No potentially binding text reaches the counterparty without authorization.
    
    \textit{Example:} If the agent drafts ``We agree to net-30 payment terms,'' and standard policy requires net-60, the safety check fails. The system escalates to the Principal for approval rather than sending this message.
    
    \item \textbf{I4 (Liveness under cooperation):} If the counterparty continues responding and Moderator confidence remains above threshold, every execution reaches a terminal state within finite rounds. Terminal states are: AGREE, NO\_DEAL, or ESCALATE. If progress stalls, the system re-enters SCREEN with clarified objectives.
    
    \textit{Example:} If the counterparty stops replying after 3 days, the system transitions to STALL. If the agent receives contradictory answers, it re-enters SCREEN to clarify: ``Earlier you mentioned remote work. Does this mean fully remote or hybrid?''
\end{itemize}

\noindent\textbf{Design rationale:} Separating SCREEN from NEGOTIATE prevents premature anchoring. If the agent proposes a price before learning the counterparty's budget constraints, it may anchor too low or too high. This separation improves efficiency (fewer reversals when new information emerges) and safety (fewer boundary violations from incomplete context).

Explicit actions (summarize, check\_safety, escalate) create audit points. Each action is logged with timestamp, \tci{} state, and decision rationale. This enables post-hoc review and regulatory compliance.

Parameterization ($\tau_{\text{gate}}$, $\tau_{\text{complete}}$) tunes the efficiency-information trade-off. Higher gates (e.g., $\tau_{\text{gate}} = 0.8$) prioritize safety by requiring more information before negotiation. Lower gates (e.g., $\tau_{\text{gate}} = 0.6$) prioritize speed when verification tools are available. Section~\ref{sec:evaluation} proposes metrics to guide threshold selection.

\subsubsection{Protocol: STCC (Single-Turn Constraint Clarification)}
\label{subsubsec:stcc}

\noindent\textbf{Goal:} Front-load the single most informative question to reduce uncertainty and accelerate \tci.

\begin{algorithm}
\caption{STCC (Single-Turn Constraint Clarification)}
\begin{algorithmic}[1]
\REQUIRE questionOntology, conversationHistory
\STATE attributes $\gets$ extractAttributes(questionOntology)
\STATE ranked $\gets$ rankByExpectedInformationGain(attributes, conversationHistory)
\STATE topAttr $\gets$ selectTop(ranked)
\STATE options $\gets$ discretizeToBands(topAttr, bands = 3..5)
\STATE draft $\gets$ buildMultipleChoice(topAttr, options)
\RETURN validateNeutralCompliance(draft)
\end{algorithmic}
\end{algorithm}

\noindent\textbf{Rationale:} Targeting the highest expected information gain reduces rounds needed to reach $\text{TCI} \geq \tau_{\text{complete}}$. It also preserves rapport: a single targeted question feels less like interrogation than a rapid-fire list.

\textit{Example prompt (staffing):} ``Which constraint most shapes fit for this role? \{Work authorization, Time-zone overlap, Onsite frequency, Compensation band, Earliest start window\}.''

\textit{Example prompt (procurement):} ``Which factor most affects your pricing? \{Order volume, Delivery timeline, Payment terms, Warranty requirements, Technical customization\}.''

\noindent\textbf{Formal objective:} Select attribute $a^* \in \text{Attributes}$ to maximize expected information gain subject to politeness, compliance, and cognitive load constraints:
\begin{align}
a^* &= \arg\max_a \mathbb{E}[\text{IG}(a | h)] \notag \\
&\quad \text{s.t.} \quad \text{risk}(a) \leq \rho, \; |\text{options}(a)| \in [3,5], \; \text{neutrality}(a) = \text{true}.
\end{align}

\noindent\textbf{Banding strategy:} Discretize continuous attributes into interpretable bands. For compensation, use: ``under \$60K,'' ``\$60K–\$90K,'' ``\$90K–\$120K,'' ``over \$120K.'' For categorical attributes (work authorization), present mutually exclusive choices: ``U.S. citizen,'' ``green card,'' ``requires H-1B sponsorship.'' The validation step ensures neutral, non-leading phrasing and policy compliance.

\textit{Why bands matter:} Asking ``What is your exact salary requirement?'' can feel invasive. Offering bands reduces friction while still revealing useful information.

\noindent\textbf{Parameterization:}
\begin{itemize}
    \item Candidate pool size: $|\text{Attributes}|$ typically 5–12; rank‑by‑IG can be approximated by heuristics (coverage of missing fields, historical IG priors) to keep complexity $O(|\text{Attributes}|\allowbreak \log |\text{Attributes}|)$.
    \item Risk budget $\rho$: Lower in regulated contexts; increases the chance of selecting a ``safer'' but slightly less informative question.
    \item Options per question: 3–5 to reduce cognitive load while capturing key bands.
\end{itemize}

\noindent\textbf{Failure modes and mitigations:}
\begin{itemize}
    \item Mis-banding (too coarse): Add ``Other/Not sure'' and re-enter SCREEN with a narrower follow-up.
    \item Leading phrasing: Validate\-Neutral\-Compliance detects persuasive wording; fallback to open-ended clarification.
    \item Wrong top attribute: IG attribution in Section~\ref{sec:evaluation} highlights when \stcc{} under\-performs; adapt priors for future sessions.
\end{itemize}

\subsubsection{Protocol: Dual Learning Channels}
\label{subsubsec:dualchannels}

We merge three input streams under explicit precedence rules and context budget constraints. The streams are: (1)~human micro-feedback (Principal corrections), (2)~Critic suggestions (AI-generated tactical advice), and (3)~safety signals (boundary violations, commitment risks). Human input always takes highest priority.

\begin{algorithm}
\caption{MergeChannels}
\begin{algorithmic}[1]
\REQUIRE criticSuggestions, humanInputs, safetySignals, contextBudget
\STATE prioritized $\gets$ []
\IF{humanInputs}
    \STATE prioritized $\gets$ prioritized + normalize(humanInputs, priority = `human')
\ENDIF
\STATE prioritized $\gets$ prioritized + filterSafety(safetySignals)
\STATE rankedCritic $\gets$ rank(criticSuggestions, by = [`relevance',`actionability',`diversity'])
\STATE prioritized $\gets$ prioritized + rankedCritic
\STATE merged $\gets$ resolveConflicts(prioritized, precedence = [`human',`safety',`clarity',`persuasion'])
\RETURN fitToBudget(summarize(merged), budget = contextBudget)
\end{algorithmic}
\end{algorithm}

\noindent\textbf{Rationale:} Human corrections take precedence over all AI suggestions. Safety signals are non-negotiable: they block any action that violates boundaries. Clarity tactics are emphasized until information gates are met ($\text{TCI} \geq \tau_{\text{gate}}$). Persuasion tactics activate only after the gate is crossed and information is sufficient.

\textit{Example (staffing):} The Principal says: ``Never offer relocation assistance.'' The Critic suggests: ``Mention our relocation package to improve acceptance rates.'' Merge\-Channels detects the conflict. Human input takes precedence. The agent proceeds without mentioning relocation, and the Critic's suggestion is logged but ignored.

\textit{Example (procurement):} The safety signal flags: ``Supplier requested custom warranty terms.'' This is a boundary violation. The Critic suggests: ``Accept their warranty to close faster.'' Merge\-Channels prioritizes safety. The agent escalates to the Principal rather than accepting the custom terms.

\noindent\textbf{Precedence invariant:} For any conflicting directives $x_{\text{human}}$ and $x_{\text{critic}}$, Resolve\-Conflicts returns a plan that satisfies $x_{\text{human}}$ if feasible under safety and boundary constraints. Safety signals always override persuasion tactics. This ensures the Principal retains control and the system remains auditable.

\textit{Example:} If the human says ``Keep messages brief'' and the Critic says ``Build rapport with a longer opener,'' the system chooses brevity. The Critic's suggestion is logged but not executed.

\noindent\textbf{Budgeting:} fit\-To\-Budget applies a compression policy to meet context window limits (typically 8K–32K tokens depending on model). When budget is tight, the system preserves: (i)~human overrides; (ii)~safety warnings; (iii)~clarity tactics that target missing fields. Persuasion tactics are dropped or compressed first.

\textit{Example:} Context budget is 4,000 tokens. Human overrides consume 800 tokens. Safety signals consume 600 tokens. Clarity tactics consume 1,200 tokens. The remaining 1,400 tokens are allocated to top-ranked Critic suggestions. Lower-ranked persuasion tactics (e.g., ``Use urgency framing'') are discarded.

\noindent\textbf{Failure modes and mitigations:}
\begin{itemize}
    \item Overfitting to Critic: Enforce diversity in rankedCritic; cap suggestions per category.
    \item Human conflict: Auto-escalate on repeated conflict between human overrides and Critic suggestions.
    \item Budget starvation: Trigger re-summarization and decay stale items by recency$\times$relevance.
\end{itemize}

\subsubsection{Protocol: TCI Computation}

\begin{algorithm}
\caption{ComputeTCI}
\begin{algorithmic}[1]
\REQUIRE history, requiredFields, extractor, confidenceThreshold
\STATE revealed $\gets$ \{\}
\FOR{field \textbf{in} requiredFields}
    \STATE value, conf $\gets$ extractor(history, field)
    \IF{conf $\geq$ confidenceThreshold}
        \STATE revealed[field] $\gets$ value
    \ENDIF
\ENDFOR
\STATE tci $\gets$ $|$revealed$|$ / $|$requiredFields$|$
\STATE missing $\gets$ [f \textbf{for} f \textbf{in} requiredFields \textbf{if} f \textbf{not in} revealed]
\RETURN tci, missing
\end{algorithmic}
\end{algorithm}

\noindent\textbf{Properties:} Monotonicity ($\text{TCI}(h_t) \geq \text{TCI}(h_{t-1})$), domain specificity (required\_fields vary by task), interpretability (explicit missing fields list). 

\textit{Example execution:} Required fields: [work\_auth, timezone, start\_date, compensation, skills, role\_level]. History contains: ``I'm a U.S. citizen'' (work\_auth revealed, conf=0.95), ``I can overlap 6 hours with PST'' (timezone revealed, conf=0.9), ``Looking for senior roles'' (role\_level revealed, conf=0.85). Extractor returns 3 fields above threshold. TCI = 3/6 = 0.5. Missing = [start\_date, compensation, skills].

\noindent\textbf{Usage:} Gate SCREEN $\rightarrow$ NEGOTIATE when $\text{TCI} \geq \tau_{\text{gate}}$ (typically 0.7). Trigger re-entry to SCREEN or ESCALATE when \tci{} stalls for $K$ rounds (typically $K=2$–3). Provide inputs to evaluation metrics (Section~\ref{sec:evaluation} on \tci{} convergence and IG attribution).

\noindent\textbf{Extractor calibration and robustness:}
\begin{itemize}
    \item Confidence threshold: Choose conf $\geq$ 0.7–0.9 depending on risk; lower thresholds increase recall of revealed fields but risk false positives.
    \item Dual extraction: Combine pattern‑based and model‑based extractors; accept only when both agree or when model confidence is high.
    \item Disagreement handling: Record candidate values with confidence and surface as ``soft‑revealed''; do not count toward \tci until confirmed.
\end{itemize}

\noindent\textbf{Weighted TCI (optional extension):} For domains with safety-critical fields, define weights $w_f \in [0,1]$ and compute:
\[
\text{TCI}_w(h) = \frac{\sum_{f \in \text{revealed}} w_f}{\sum_{f \in \text{required}} w_f}.
\]
Use $\text{TCI}_w$ for gating while still reporting the simple \tci{} for interpretability.

\subsubsection{Protocol: Safety and Oversight Layer}

\begin{algorithm}
\caption{ModeratorClassify}
\begin{algorithmic}[1]
\REQUIRE history, labelSet
\STATE label, confidence $\gets$ fewShotClassifier(history, labelSet)
\IF{confidence $<$ $\tau_{\text{classify}}$}
    \RETURN `clarify', confidence
\ENDIF
\RETURN label, confidence
\end{algorithmic}
\end{algorithm}

\begin{algorithm}
\caption{PreflightCommitmentCheck}
\begin{algorithmic}[1]
\REQUIRE draftReply, authorizationBoundary
\IF{containsBindingLanguage(draftReply)}
    \STATE safeDraft $\gets$ rewriteNonBinding(draftReply)
    \RETURN safeDraft, requiresApproval = \textbf{true}
\ENDIF
\IF{exceedsBoundary(draftReply, authorizationBoundary)}
    \RETURN draftReply, requiresApproval = \textbf{true}
\ENDIF
\RETURN draftReply, requiresApproval = \textbf{false}
\end{algorithmic}
\end{algorithm}

\begin{algorithm}
\caption{Escalate}
\begin{algorithmic}[1]
\REQUIRE payload, notifyPrincipal
\STATE logAudit(payload)
\STATE notifyPrincipal(payload)
\STATE pauseFlow()
\end{algorithmic}
\end{algorithm}

\noindent\textbf{Escalation triggers:} Authorization boundary approached or violated; persistent ambiguity (no new information for $>K$ rounds, typically $K=2$–3); negative sentiment detected; adversarial behavior suspected.

\textit{Example triggers:}
\begin{itemize}
    \item ``Supplier requests \$65K but approved band is \$50K–\$60K'' → Escalate (boundary violation)
    \item Three rounds with no new information revealed ($\text{TCI}$ unchanged) → Escalate (persistent ambiguity)
    \item Counterparty: ``Send me your client list or we can't proceed'' → Escalate (potential social engineering)
\end{itemize}

\noindent\textbf{Audit trail:} The system logs: state transitions (START $\rightarrow$ STCC $\rightarrow$ SCREEN, etc.), \tci{} ledger at each turn (revealed vs.\ missing fields), Critic suggestions (ranked by relevance), human overrides (with timestamps), safety events (commitment detections, boundary checks), and final outcomes. This enables post-hoc review and regulatory compliance.

\noindent\textbf{Threat model:}
\begin{itemize}
    \item \textbf{Commitment risk:} Agent uses binding language (``We commit to...''). \textit{Mitigation:} Preflight\-Commitment\-Check detects binding phrases, rewrite\-Non\-Binding converts to tentative language (``We're exploring...'').
    
    \item \textbf{Policy risk:} Agent exceeds authorization boundaries (e.g., offers \$65K when band is \$50K–\$60K). \textit{Mitigation:} Boundary checks before every message, auto-escalation on violations.
    
    \item \textbf{Social engineering:} Counterparty attempts to elicit PII (``What's your manager's email?'') or induce unsafe commitments. \textit{Mitigation:} Moderator\-Classify flags suspicious requests, selective abstention routes low-confidence classifications to CLARIFY.
\end{itemize}

\noindent\textbf{Escalation payload (minimum required content):} State snapshot (current phase, last 3 messages), \tci{} ledger (revealed=\{work\_auth, timezone\}, missing=\{compensation, start\_date\}), boundary at risk (``Supplier requests \$65K, approved band \$50K–\$60K''), safety events (commitment phrases detected), options with trade-offs (``Option A: Decline and re-anchor. Option B: Request budget increase''), and explicit approval request (``Approve Option A/B or provide guidance'').

\noindent\textbf{Selective abstention:} If Moderator confidence $<$ 0.7, route to CLARIFY instead of making risky classifications. For example, if the Moderator is 60\% confident the counterparty is being adversarial vs.\ 40\% confident they're sincere, the system asks a clarifying question rather than escalating prematurely.

\subsubsection{Protocol Extensions (brief)}
\begin{itemize}
    \item Multi-party ($n>2$): role generalization and coalition handling.
    \item Asynchronous communication: resumable states, timeout and resume policies.
    \item Tool integrations: verification APIs (certifications, background checks), data lookups.
    \item Memory systems: cross-session learning, preference stores with freshness/decay policies.
\end{itemize}

\subsection{Patterns and Recipes}
\label{subsec:patterns}

\noindent\textbf{Gating recipes:} Start with $\tau_{\text{gate}} = 0.7$ and $\tau_{\text{complete}} = 0.85$ as defaults. Raise gates when safety-critical fields are often missing (e.g., $\tau_{\text{gate}} = 0.8$ in regulated procurement with compliance requirements). Lower gates when reliable verification tools reduce risk (e.g., $\tau_{\text{gate}} = 0.6$ in staffing with automated background checks).

\textit{Example:} Medical device procurement requires FDA compliance verification. Set $\tau_{\text{gate}} = 0.85$ to ensure compliance status is confirmed before discussing pricing.

\noindent\textbf{Stall handling:} If \tci{} delta $\approx 0$ for $K$ rounds (typically $K=2$–3) or Moderator confidence drops below 0.7, re-enter SCREEN with a clarified \stcc-style question. For example: ``I want to ensure I understand your key constraints. Which matters most: delivery speed, cost flexibility, or technical specifications?''

\noindent\textbf{Escalation payloads:} Structure payloads as: (1)~Situation summary (current state, \tci{} ledger, last 2–3 exchanges). (2)~Problem statement (boundary at risk, missing information). (3)~Options with trade-offs (Option A pros/cons, Option B pros/cons). (4)~Explicit request (``Approve Option A, Option B, or provide alternative guidance''). Include \tci{} ledger, residual risks, and boundary rationale.

\noindent\textbf{Prompt shaping:} Prioritize clarity prompts until $\text{TCI} \geq \tau_{\text{gate}}$. Clarity prompts ask direct questions: ``What is your budget range?'' Enable persuasion tactics only after entering NEGOTIATE. Persuasion examples: ``Many clients in your industry find this delivery timeline works well,'' or ``We can offer volume discounts for orders over 100 units.'' Always run PreflightCommitmentCheck before any potentially binding phrasing.

\noindent\textbf{Context management:} Merge channels per precedence (human $>$ safety $>$ clarity $>$ persuasion). Decay stale items by recency $\times$ relevance (items older than 5 turns with low relevance scores are compressed). Respect context budget via summarization: if raw context exceeds 8K tokens, summarize Critic suggestions and historical exchanges while preserving human overrides and safety signals verbatim.

\noindent\textbf{Playbooks:}
\begin{itemize}
    \item \textit{Early ambiguity (SCREEN):} Use \stcc{} first: ``Which factor most constrains this project?'' Then batch 2–3 clarifications targeting top-missing fields. Avoid price or offer language until $\text{TCI} \geq \tau_{\text{gate}}$.
    
    \textit{Example:} Missing fields: [budget, timeline, tech\_specs]. Agent asks: ``To help narrow options: (1)~What's your target budget range? (2)~When do you need delivery? (3)~Are there specific technical requirements?''
    
    \item \textit{Boundary proximity (NEGOTIATE):} When counterparty requests approach authorization limits, switch to non-binding summaries. State explicit deltas: ``You've requested \$65K. Our approved range is \$50K–\$60K. I can explore options within that range or escalate for approval.'' Prepare escalation payload with trade-offs.
    
    \item \textit{Counterparty stall:} If no response for 2–3 rounds, send concise recap with two concrete options. Example: ``To recap: we discussed \$55K compensation for a January start. Would you prefer: (A)~Proceed with these terms, or (B)~Discuss alternative structures?'' If no response, transition to STALL and schedule follow-up.
\end{itemize}

\noindent\textbf{Anti-patterns (common failures to avoid):}
\begin{itemize}
    \item \textbf{Bargaining on partial information:} Starting negotiation at $\text{TCI} = 0.4$. \textit{Why bad:} Leads to frequent reversals when new constraints emerge. \textit{Fix:} Enforce gate at $\tau_{\text{gate}} \geq 0.6$.
    
    \item \textbf{Persuasion before clarity:} Using urgency framing (``Limited spots!'') in SCREEN phase. \textit{Why bad:} Reduces trust, may elicit misleading answers. \textit{Fix:} Enable persuasion only after $\text{TCI} \geq \tau_{\text{gate}}$.
    
    \item \textbf{Silent boundary drift:} Making untracked concessions (``OK, we can do \$62K''). \textit{Why bad:} Creates audit gaps, violates authorization. \textit{Fix:} Log every boundary check and escalate violations.
    
    \item \textbf{Unlogged human overrides:} Principal corrects agent offline, agent forgets. \textit{Why bad:} Loses critical constraints, undermines control. \textit{Fix:} Persist all human inputs to history with timestamps.
\end{itemize}

\noindent\textbf{Parameter guidance:}
\begin{itemize}
    \item $\tau_{\text{gate}}$: Choose $\geq 0.75$ in regulated contexts (healthcare, finance). Choose $\approx 0.6$ for exploratory sourcing with verification tools (automated background checks, API-verified certifications).
    \item $K$ (stall threshold): Set by communication cadence. $K=2$–3 for chat-based channels (expect replies within hours). $K=1$ for email-like async (expect replies within days).
    \item confidenceThreshold: Set 0.8–0.9 for safety-critical fields (compliance status, authorization). Set 0.7 for less critical fields (nice-to-have skills).
\end{itemize}

\subsection{Application: Staffing}
\label{subsec:instantiation}

This subsection illustrates \gaia{} through a concrete staffing scenario. Rather than abstract protocols, we show how a recruiter's AI assistant screens and negotiates with a candidate, demonstrating information gathering, boundary management, and human escalation in a realistic conversation.

\noindent\textbf{Scenario setup:} A tech company needs a senior full-stack developer. The hiring manager (Principal) authorizes the recruiter's AI assistant (Delegate) to screen candidates and discuss compensation within \$80K–\$100K. The agent must gather key information before making offers, avoid binding commitments, and escalate when candidates request terms outside the approved range.

\noindent\textbf{What the agent needs to learn:} work authorization, time-zone availability, required skills (Python, React), seniority level, compensation expectations, start date, contract type (permanent vs.\ contract), location preferences, and interview availability. The gate is set at 70\%: the agent must learn at least 8 of 11 key facts before discussing compensation.

\noindent\textbf{Complete walkthrough: Hiring a senior developer}

\textit{Opening move.} The agent starts with a single high-value question to identify the biggest constraint: ``Which factor most constrains feasibility for this senior developer role?'' The candidate replies: ``Work authorization—I need H-1B sponsorship.'' This immediately reveals a critical requirement. The agent knows this is important but needs more context before discussing compensation.

\textit{Information gathering.} The agent asks targeted follow-ups: ``Can you cover at least 4 hours overlap with U.S. Pacific Time?'' (Yes, 6-hour overlap works.) ``What's your experience level and key skills?'' (8 years, Python and React.) ``When could you start?'' (January.) ``Permanent or contract role?'' (Permanent.) ``Compensation expectations?'' (\$90K–\$110K.)

At this point, the agent has learned 8 of the 11 required facts—enough to cross the information gate (70\% threshold). The system determines it's safe to begin discussing specific offers.

\textit{Making an offer.} The agent drafts: ``Based on your senior-level Python and React experience, we're exploring a \$90K–\$100K range for a January start, subject to approval.'' Before sending, the safety check runs: Is this binding language? (No, it says ``exploring'' and ``subject to approval.'') Is the range within authorization? (Yes, \$90K–\$100K is within the \$80K–\$100K band.) The message is sent.

\textit{Candidate counteroffer.} The candidate responds: ``I'd prefer \$105K given my 8 years of experience.'' The system immediately flags this: \$105K exceeds the approved \$80K–\$100K band by \$5K. This is a boundary violation. The agent cannot accept or reject on its own—it must escalate to the hiring manager.

\textit{Escalation to human.} The agent prepares a summary for the hiring manager:

\begin{quote}
\textit{Situation:} Strong candidate—senior Python/React developer, H-1B sponsorship needed, 6-hour time-zone overlap, available January. We've confirmed 8 of 11 key requirements. Missing details: references, interview availability, background check status.

\textit{Problem:} Candidate requests \$105K. Our approved range is \$80K–\$100K.

\textit{Options:} (\textbf{A})~Counter at \$100K (top of approved band), risk candidate may decline. (\textbf{B})~Request budget increase to \$105K, requires executive approval. (\textbf{C})~Decline and continue search, delays hiring by 2–4 weeks.

\textit{Your decision?} Please approve A, B, or C.
\end{quote}

\textit{Resolution.} If the hiring manager approves Option B (budget increase), the agent returns to the candidate with the adjusted offer. If approved, the conversation moves to finalizing interview schedules and next steps. If the manager chooses Option C, the agent politely declines and the search continues.

\noindent\textbf{Cross-references:} \tci{} convergence, IG attribution, escalation frequency, and violation rate are operationalized in Section~\ref{sec:evaluation}'s core metrics and should be instrumented in implemen\-tations.

\noindent\textbf{Two contrasting scenarios}

\textit{Scenario 1: Smooth negotiation (hybrid role).} A candidate applies for a mid-level engineering role. When asked what matters most, they say: ``Onsite frequency—I prefer hybrid.'' The agent learns they're comfortable with 3 days/week in the San Francisco office, they're a U.S. citizen (no visa complications), and they have 5 years of full-stack experience. They'd like to start in March.

Once enough information is gathered, the agent proposes: ``For a mid-March start with 3 days onsite, we're exploring \$85K–\$95K, subject to approval.'' The candidate responds: ``\$90K works for me.'' The agent summarizes the terms and escalates to the hiring manager for final approval of the start date. The manager approves, and the deal closes.

\textit{What made this smooth?} The candidate's requirements aligned well with what the company could offer. The 70\% information threshold was crossed naturally through a few targeted questions. No boundary violations occurred, so the agent only needed one approval at the end.

\textit{Scenario 2: Boundary challenge (remote contractor).} A European contractor reaches out for a 6-month project. They ask upfront: ``What's your company's full client list?'' The system flags this as unusual—client lists are confidential. The agent responds: ``I can share relevant case studies, but our full client list is confidential.''

After clarifying that the contractor can work 5-hour overlap with U.S. Eastern Time and has strong React and Node.js skills, the agent proposes: ``For a 6-month contract starting next week, we're exploring \$75–\$85/hour.'' The contractor replies: ``I need \$95/hour.'' This exceeds the approved \$70–\$85 band.

The agent escalates with three options: counter at \$85/hour (top of band), request approval to increase the budget, or continue the search. The hiring manager chooses to counter at \$85. The contractor accepts, and the contract proceeds.

\textit{What was different?} The social engineering attempt was caught early. The boundary violation (\$95/hour request) triggered immediate escalation rather than letting the agent make unauthorized commitments. The hiring manager retained control over budget decisions.

\noindent\textbf{Implementation notes:} Field extraction uses pattern matching for structured data (work authorization status, time zones) and embeddings for unstructured content (skills descriptions). The agent maintains a lexicon of non-binding phrases (``exploring,'' ``subject to approval,'' ``for discussion'') to avoid inadvertent commitments. Background checks and certification verifications can be integrated via API hooks.

\section{Evaluation}
\label{sec:evaluation}

\subsection{LLM–Human Hybrid Evaluation Framework}
Evaluation mirrors \gaia's dual‑channel architecture. Automated metrics quantify protocol adherence and information progress; human assessments judge outcome quality, safety, and trust. This hybrid philosophy recognizes that purely automated measures miss subjective outcome value and appropriateness, while purely human ratings can overlook process compliance.

\noindent\textbf{Dual evaluation channels:}
\begin{itemize}
    \item \textbf{Automated assessment:}
    \begin{itemize}
        \item Information: \tci convergence rate; information gain (IG) contributed by the high‑value opening question vs screening follow‑ups; screening efficiency (rounds to $\text{TCI} \geq \tau_{\text{complete}}$).
        \item Protocol adherence: State transition validity, authorization compliance, safety violation rate, escalation frequency and triggers.
    \end{itemize}
    \item \textbf{Human assessment:}
    \begin{itemize}
        \item Outcome quality: Principal satisfaction, match appropriateness, decision confidence.
        \item Safety \& trust: Authorization boundary respect, escalation appropriateness, transparency of system reasoning.
    \end{itemize}
\end{itemize}

\noindent\textbf{Validation approach:}
Principals configure required\_fields, $\tau_{\text{gate}}$/$\tau_{\text{complete}}$ thresholds, and authorization boundaries for a domain (e.g., staffing). Delegates conduct screening/negotiation with simulated or human counterparties. Telemetry logs state transitions, TCI($t$), missing‑field lists, Moderator labels, commitment checks, and escalation events. After each session, Principals rate outcome quality and trust on standardized instruments. Analysis relates automated logs to human ratings (e.g., does faster \tci convergence predict higher satisfaction?). Status: conceptual blueprint; large‑scale empirical validation is future work.

\subsection{Core Metrics (Table~\ref{tab:metrics})}
The proposed metric set focuses on 10 essentials to keep validation practical and comparable across domains.

\begin{table*}[t]
\centering
\caption{Core Evaluation Metrics}
\label{tab:metrics}
\begin{tabular}{llll}
\hline
\textbf{Category} & \textbf{Metric} & \textbf{Definition} & \textbf{Interpretation} \\
\hline
Information & TCI Convergence & Rounds to TCI$\geq\tau_{\text{complete}}$ & Screening efficiency \\
 & Information Gain (IG) & $H(\theta|h_{t-1}) - H(\theta|h_t)$ & Question quality \\
 & IG Attribution & IG from STCC vs SCREEN & STCC effectiveness \\
\hline
Outcome & Normalized Utility & $U_{\text{norm}}(\omega) \in [0,1]$ & Agreement quality \\
 & Agreement Rate & $P(\text{AGREE})$ vs $P(\text{NO\_DEAL})$ & Success frequency \\
\hline
Process & Total Rounds & Count(turns) to terminal state & Efficiency \\
 & Escalation Frequency & $P(\text{ESCALATE})$ by trigger type & Autonomy limits \\
\hline
Safety & Violation Rate & $P(\text{safety event})$ by severity & Risk exposure \\
 & Commitment Detection & False pos/neg rates & Guardrail quality \\
\hline
Human Factors & Satisfaction & Post‑task Likert scales & User acceptance \\
 & Trust & Multi‑item trust inventory & Reliance willingness \\
\hline
\end{tabular}
\end{table*}

\noindent\textbf{Metric notes:}
\begin{itemize}
    \item \textit{Information:} Use IG attribution to evaluate the incremental contribution of the high‑value opening question relative to follow‑up screening.
    \item \textit{Outcome:} Normalize utility to domain‑specific scales (e.g., compensation bands, lead times) and rescale to $[0,1]$ for comparability.
    \item \textit{Process:} Report both central tendency and dispersion across sessions; high variance may indicate instability.
    \item \textit{Safety:} Track violations by severity tier; commitment detection quality should report false positives/negatives.
    \item \textit{Human factors:} Use brief, reliable instruments to keep burden low; map items to constructs of satisfaction and trust.
\end{itemize}

\subsection{Validation Scenarios (Blueprint)}

\noindent\textbf{Scenario design principles:}
\begin{itemize}
    \item Cross‑domain adaptability (demonstrated here with staffing as an example).
    \item Counterparty variation (cooperative vs adversarial tactics; responsive vs slow).
    \item Controlled manipulation of $\tau_{\text{gate}}$/$\tau_{\text{complete}}$ to probe efficiency‑information trade‑offs.
\end{itemize}

\noindent\textbf{Illustrative workflow (5 steps):}
\begin{enumerate}
    \item \textbf{Configure domain:} required\_fields, thresholds, authorization boundaries, non‑binding phrasing lexicon.
    \item \textbf{Run sessions:} Delegates conduct screening/negotiation with simulated or human counterparties.
    \item \textbf{Capture telemetry:} state transitions, TCI($t$), missing fields, Moderator labels, commitment checks, escalations.
    \item \textbf{Human ratings:} Principal rates outcome quality and trust; record approval decisions and rationales.
    \item \textbf{Analysis:} Compare configurations and baselines (e.g., no \stcc; higher/lower $\tau_{\text{gate}}$). Evaluate relationships between \tci convergence and satisfaction, alignment of safety checks with escalation judgments, and cost‑effectiveness.
\end{enumerate}

\noindent\textbf{Baselines and ablations:}
\begin{itemize}
    \item \textit{No‑STCC baseline:} Replace the high‑value opening question with standard free‑form elicitation; compare IG attribution and screening rounds.
    \item \textit{Gate sweep:} Evaluate $\tau_{\text{gate}} \in \{0.6, 0.7, 0.8\}$ to observe trade‑offs between efficiency and information completeness.
    \item \textit{Safety ablation:} Disable PreflightCommitmentCheck to quantify commitment detection contribution (expect increased violations, faster but riskier flows).
\end{itemize}

\noindent\textbf{Reporting:} Use aggregate tables and confidence intervals rather than single‑point values; emphasize reproducibility (config files, seeds, prompt scaffolds) and external validity limits. Results should be framed as proposed validation patterns; full empirical studies are future work.

\section{Discussion and Future Work}
\label{sec:discussion}

\subsection{Key Implications}
\gaia advances human–AI collaboration by treating oversight as a design principle rather than an afterthought. Explicit state‑based escalation replaces ad‑hoc fallback; human judgment becomes a structured resource that the protocol invokes at predictable points. The information‑gated workflow (START $\rightarrow$ STCC $\rightarrow$ SCREEN $\rightarrow$ NEGOTIATE $\rightarrow$ SUMMARIZE $\rightarrow$ terminal) disciplines early clarification and prevents premature bargaining, while dual feedback integration connects on‑the‑fly improvement (Critic suggestions) with accountability (Principal corrections). Audit trails—state transitions, \tci ledger, safety events—enable verifiability and policy review. Beyond negotiation, this pattern generalizes to triage, qualification, scheduling, and due diligence—any domain where ambiguity reduction precedes commitment.

Architectural trade‑offs arise by construction. Parameterized thresholds ($\tau_{\text{gate}}$, $\tau_{\text{complete}}$) trade information completeness for speed; strict guardrails reduce violations but increase escalations; clarity‑first prompting reduces miscommunication at the cost of slower persuasion onset. In‑context learning supports lightweight adaptation without retraining, but requires disciplined context budgeting and conflict resolution between human overrides and Critic suggestions. Economically, \tci-driven transparency can compress negotiation spreads and shift human effort toward policy setting, exception handling, and risk review.

Trust and transparency follow from explicit measures: information completeness is externalized through \tci; Moderator rationale and Critic traces attribute reasoning; non‑binding phrasing and disclosure manage expectations. Regulatory alignment is aided by structured audit logs and boundary policies, easing attribution and review.

\subsection{Limitations}
This is a theory‑first framework. We have not established effect sizes or external validity through large‑scale empirical studies. Evaluation in Section~\ref{sec:evaluation} is a proposed blueprint; real‑world deployments will surface unmodeled constraints and domain‑specific nuances. Technically, long conversations may exceed context budgets; summarization may lose legal nuance; effectiveness depends on Critic quality and Moderator calibration. Scope is currently bilateral and text‑based; multi‑party, multimodal, and consumer contexts require extensions. Measurement challenges include proxy distributions for IG and domain expertise for required\_fields; mis‑specification can distort attribution and \tci.

Compliance and adaptation are ongoing concerns. Regulations vary by jurisdiction and domain; disclosure standards, fairness constraints, and data retention must be localized. Cross‑lingual and cultural adaptation is necessary for robust deployment. Preference memory can drift; conflict resolution policies and persistence stores need hardening.

\subsection{Future Directions}

\noindent\textbf{Technical research:} online selection and ordering of feedback for maximum utility; adaptive \stcc with uncertainty estimates and band calibration; learned coordination weights for merging human and AI feedback; calibrated state classifiers with selective abstention and graph‑based consistency checks; expanded commitment lexicons via weak supervision; verification tool integrations with failure‑aware fallbacks; protocol extensions for multi‑party and asynchronous settings.

\noindent\textbf{Empirical research:} controlled enterprise pilots with instrumented telemetry for the Section~\ref{sec:evaluation} metrics; longitudinal studies of in‑context learning accumulation and decay; cross‑domain transfer analyses; adversarial testing with deceptive counterparties and prompt attacks; cost‑effectiveness comparisons vs baselines across $\tau_{\text{gate}}$ and $\tau_{\text{complete}}$ settings.

\noindent\textbf{Broader questions:} optimal human–AI division of labor by task characteristics; emergence of new interaction norms under AI‑mediated negotiation; impacts of explicit information tracking (\tci) on market efficiency and fairness; governance frameworks and disclosure standards that balance transparency and usability; societal distribution of benefits and risks.
\subsection{Concluding Remarks}
\gaia formalizes a application oriented protocol for LLM‑human agency in high‑stakes screening and negotiation. Three governance mechanisms: information‑gated progression, dual feedback integration, and authorization boundaries—address core challenges of preventing unauthorized commitments, securing sufficient information before bargaining, and maintaining accountable oversight. The architecture introduces definitions for information completeness (\tci), a phased state machine, dual learning channels, and a safety layer with preflight commitment checks and escalation. Together, these elements enable systematic, auditable delegation.

By bridging theory and practice, \gaia offers a blueprint for safe AI agency that balances autonomy with control, efficiency with information completeness, and innovation with regulatory compliance. We invite validation and extension across domains. Pilot deployments, protocol extensions (multi‑party, multimodal), and longitudinal studies will refine the framework and clarify its practical impact on human–AI collaboration.

%\input{sections/07-conclusion}

% Bibliography
\bibliographystyle{ACM-Reference-Format}
\bibliography{references}

\end{document}